# Multilayer Graph-Based Trajectory Planning for Race Vehicles in Dynamic Scenarios

Tim Stahl[1,3], Alexander Wischnewski[2], Johannes Betz[1], Markus Lienkamp[1]

*Abstract*— Trajectory planning at high velocities and at the handling limits is a challenging task. In order to cope with the requirements of a race scenario, we propose a far-sighted two step, multi-layered graph-based trajectory planner, capable to run with speeds up to 212 km/h. The planner is designed to generate an action set of multiple drivable trajectories, allowing an adjacent behavior planner to pick the most appropriate action for the global state in the scene. This method serves objectives such as race line tracking, following, stopping, overtaking and a velocity profile which enables a handling of the vehicle at the limit of friction. Thereby, it provides a high update rate, a far planning horizon and solutions to non-convex scenarios. The capabilities of the proposed method are demonstrated in simulation and on a real race vehicle.

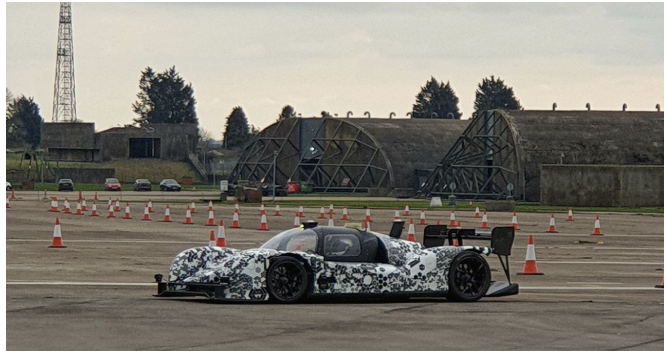

Fig. 1: *DevBot 2.0* - The development vehicle of *Roborace* holding various sensors and control units. [1]

## I. Introduction

Motion planning for autonomous vehicles has been an active research field for several decades. As the computational power improves, further requirements to be met by a motion planner can be addressed. The first mobile robots had holonomic platforms and moved at low speeds. Later, as the speed of vehicles increased, higher demands such as non-holonomic kinematic validity and heading or curvature steady paths were imposed. Finally, static and dynamic obstacles require real-time capability and online adaptability.

Roborace [1] claims to be the first autonomous race series. The team at the Technical University Munich developed a software stack autonomously driving the vehicle at a race event in 2018 [2]. While in 2018 all events where carried out in a static environment, future races are planned to include several participants sharing the track. Such a setup opposes new challenges to a local trajectory planner, which will be tackled in this paper.

The proposed approach is designed to plan trajectories in a race scenario at high speeds. On one hand, the planner must be able to follow a race optimal line, when no object is present. On the other hand, the planner must be able to react appropriately, when objects are in the vehicles vicinity. Furthermore, due to the semi-structured environment of a race track (i.e. no lane markings/specification), a high planning horizon must be achieved in order to guarantee recursive feasibility.

[1]Chair of Automotive Technology, Technical University of Munich, Munich, Germany
[2]Chair of Automatic Control, Technical University of Munich, Munich, Germany
[3]Corresponding Author, `stahl@ftm.mw.tum.de`

Our method covers a far (e.g. 200 m) planning horizon including sophisticated overtake maneuvers instead of simple lane change motions. In order to cope with the wide range of accelerations and velocities of a race scenario, the planner follows a two-step approach ensuring manageable state space dimensions. The planner has been evaluated in simulation and on a real race vehicle (Figure 1).

## II. Related Work

In the past decades, a vast variety of approaches tackling motion planning have been proposed. Recent surveys [3]–[5] identified clusters of fundamental concepts. Combining the findings of the three, the following four classes of motion planners have been identified:

- Sampling-based approaches/incremental search
- Interpolation methods
- Numerical optimization/variational methods
- Graph-based approaches

This section focuses on the latter two approaches, since *sampling-based* methods find their strengths in unstructured or unknown environments [6] and *interpolation* primarily focuses on the smooth transition between multiple coordinates instead of the actual planning task.

Within *numerical optimization* strategies, the motion planning problem is described mathematically as a minimization or maximization problem. A major difficulty lies in the tendency of these methods to detect local minima only. A common approach is to formulate the problem in a convex manner, which restricts the solution space.

Optimization-based motion planning is an active research field with a large number of publications, especially dedicated to model predictive control (MPC). Most of the research focuses on urban or highway applications, which differ often with respect to their assumptions, requirements and boundary conditions [3]–[5]. A few research projects [7]–[9] address autonomous racing with optimization methods ranging from optimal control to learning MPC. None of these approaches addresses dynamical objects, which would require longer planning horizons in semistructured environments and questions the real-time applicability.

*Graph-based* approaches span a spatial or spatiotemporal, one dimensional or hierarchical tree along the drivable area. Commonly, each node in the graph holds a cost, calculated based on identified optimality metrics. By utilizing graph-search algorithms, the path minimizing the cost between adjacent nodes can be identified. Graph-based algorithms have proven to be effective for autonomous driving [10]–[14].

Several authors [10]–[12] span a one-layered graph in the spatial dimension, which offers various lateral and sometimes longitudinal target states along the track. Some include a dynamic object handling within a short time horizon. Gu et al. [13] extend the spatial dimension to a hierarchical tree with linear edges covering the track. The cost minimal path through the graph is used as an input for a path optimization. McNaugthon et al. [14] spanned a spatiotemporal hierarchical tree.

However, to serve a race scenario, the range of velocities and acceleration profiles needed within a spatiotemporal approach (curse of dimensionality) would be too large for a real-time graph representation. Besides, one-layered graphs lack the capability to model sophisticated maneuvers, like an overtaking maneuver. This can result in critical situations when facing a race scenario, e.g. initiating an overtaking maneuver, while a sharp turn is ahead. We tackle those issues by spanning a spatial hierarchical tree with a far planning horizon and plan the temporal dimension in an adjacent step. We extend the concept of Gu et al. [13] by splitting the approach in an offline and online phase, sampling of realistic probe paths instead of linear ones and the usage of action sets.

The contribution of this paper is a trajectory planner based on a state lattice interpreted as a multilayer directed graph, designed to cope with the requirements of a semistructured race environment. In this spirit, the planner provides a high update rate (suited for high velocities, i.e. above 10 Hz), a far planning horizon (guarantee feasibility with sharp curves and without lane markings, e.g. 200 m), solutions to non-convex scenarios, and a velocity profile maxing out the friction potential along the track. The method generates multiple action primitives in realtime, which are evaluated by an adjacent behavior planning module. We demonstrate the capabilities and performance of the planner on a race vehicle, the *DevBot 2.0* (Figure 1), with velocities up to 212 km/h.

## III. Motion Planning Framework

We use a motion-planning architecture consisting of three main components: global trajectory planner, local trajectory planner and adjacent behavior planner (Figure 2). We use a global planner, proposed by Heilmeier et al. [15], which uses optimization to generate an ideal trajectory in a static environment. This trajectory is used as a reference line and velocity constraints for the local planner. The behavior planner is the last module in the chain and receives one or multiple trajectory candidates from the local planner to choose from. Therefore, the local trajectory planner establishes (whenever feasible) an action set containing fixed action primitives like "keep straight," "overtake left" and "overtake right." The behavior planner evaluates all options and selects one trajectory to be forwarded to the controller.

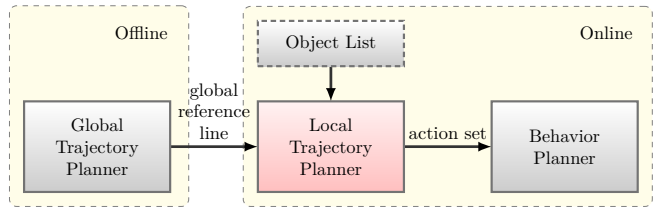

Fig. 2: Interfaces of the local trajectory planner in the motion planning stack.

In order to reduce the online computation load, our approach picks up on detailed offline computed data (Figure 3). The idea is to compute as much information as possible prior to the online planning task. In that way, the online part of the algorithm has more computational resources to solve the online planning problem.

## IV. Method

As described in Section III, the local trajectory planner is partitioned into an offline and an online part. The idea is to precompute as much information as possible in the offline part (Section IV-A) in order to reduce the computational load in the online part (Section IV-B). The following sections cover the components in more detail.

### A. Offline Part

The offline preparation can be sectioned into three steps. First, the overall lattice is laid out by the definition of vertices (Section IV-A.1). Second, splines connecting pairs of vertices are generated (Section IV-A.2). Finally, a cost is assigned to each edge in the graph (Section IV-A.3).

*1) Lattice definition:* The state lattice is generated offline and defined in the Frenét space along a reference line. This line is defined as discrete function $[x(s), y(s), \theta(s), \kappa(s)]$ along the arc length $s$, also called station. The Frenét frame is defined as the coordinate system spanned by the tangential vector $t$ and the normal vector $n$ at any point of the reference line.

Any point along the reference line is specified by the continuous arc length $s$ and a lateral displacement $l$. For every point $[s, l]$ in the Frenét space, the corresponding Cartesian coordinate $p$ is defined as

$$p(s, l) = r(s) + n(s)l, \qquad (1)$$

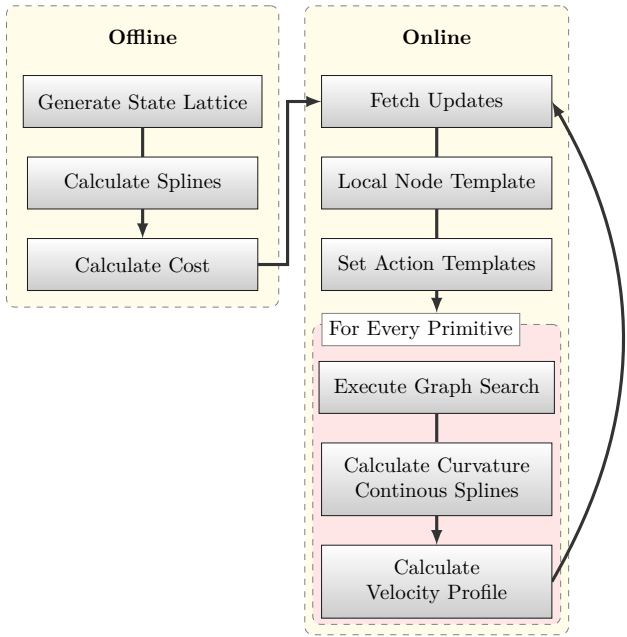

Fig. 3: Framework of the proposed local trajectory planner – decomposed into an offline and an online part.

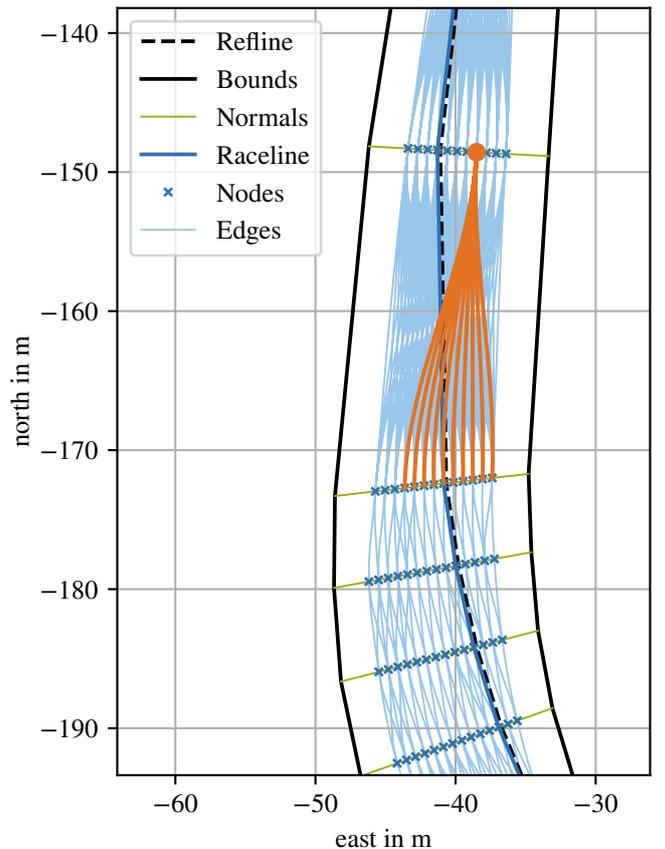

Fig. 4: Exemplary detail of a graph with a straight and beginning curve segment. The edges originating from an arbitrary node are highlighted in orange.

where $r(s)$ holds the Cartesian coordinate of the reference line at station $s$. Since one aims for fast lap times in a race scenario (especially when obstacle free), the reference line is chosen to be an (close to) optimal race line, e. g. time or curvature minimal [15].

The state lattice is formed by layers $L_i$ distributed along the station $s$. In order to guarantee proper curve tracking and smooth transitions on the straights, sections of higher curvature $\kappa$ obtain a more dense layer coverage than straight sections (Figure 4). A layer is defined as an agglomeration of vertices or nodes $n_j \in L_i$ along the normal vector at station $s_i$. The vertices of each layer are sampled with a fixed lateral displacement to the left and right relative to the reference line until the track bounds are reached. The associated heading $\theta$ of each vertex is interpolated linearly between the heading of the race line and the heading of corresponding track bound.

*2) Spline generation:* All vertices of neighboring layers, not exceeding a relative vehicle specific lateral displacement ratio, are connected with an edge, i. e. a path segment. Cubic polynomials are used to describe a $C_1$ continuous transition between each pair of poses. It should be noted that the paths are sampled in Cartesian coordinates, in order to ease up collision checking, determination of path curvature and lateral accelerations.

The course of the $x$ and $y$ coordinate is each described by a cubic polynomial

$$p(\mu) = a_{3,\mathrm{p}}\mu^3 + a_{2,\mathrm{p}}\mu^2 + a_{1,\mathrm{p}}\mu + a_{0,\mathrm{p}}, \qquad (2)$$

with the shared virtual path variable $\mu \in [0,1]$. The variable $p$ is to be substituted by $x$ and $y$ as well as $f_{\mathrm{cs}}(\cdot)$ by $\cos(\cdot)$ and $\sin(\cdot)$, respectively. The coefficients $a_i$ are chosen in a way that the constraints

$$\begin{aligned} p_s &= a_{0,\mathrm{p}} \\ p'_s &= s_{\mathrm{len}} f_{\mathrm{cs}}(\theta_s) = a_{1,\mathrm{p}} \\ p_e &= a_{3,\mathrm{p}} + a_{2,\mathrm{p}} + a_{1,\mathrm{p}} + a_{0,\mathrm{p}} \\ p'_e &= s_{\mathrm{len}} f_{\mathrm{cs}}(\theta_e) = 3a_{3,\mathrm{p}} + 2a_{2,\mathrm{p}} + a_{1,\mathrm{p}} \end{aligned} \qquad (3)$$

are fulfilled. In order to align with the range of the virtual path variable $\mu$, the heading constraint is scaled by the path length $s_{\mathrm{len}}$. Since the path length is not known beforehand, it is initially estimated by the euclidean distance of the end points and iteratively refined by calculating the spline coefficients multiple times. Since this process belongs to the offline phase, additional calculation time can be neglected.

When analyzing the constraints given in Equation 3, one can see that a curvature steady ($C_2$) course is not guaranteed. An overall curvature steady lattice would lead to a drastic expansion of the state space, since every possible pair of connected path segments would need their unique curvature transition and node. Since this would make the online application intractable, we use the $C_1$ continuous lattice for collision checks and select a cost optimal path sequence. The selected path is then online solved for $C_2$ continuity (Section IV-B). Compared to linear node transitions, the $C_1$ splines form a similar path as the $C_2$ splines and serve a reasonable collision check.

Path segments exceeding the physical abilities of the vehicle (e.g. curvature violating the turn radius of the vehicle) are not added to the graph. Furthermore, all edges ending in nodes with no children or starting from nodes with no parent are iteratively removed. Finally, parameters like spline length, course of heading and curvature are calculated for each spline.

*3) Cost function:* In order to enable an online graph search, each edge is associated with a dedicated cost value. The magnitude of the cost value describes the undesirability of the vehicle following that path. The cost $c_{i,j}$ of the segment between the nodes $i$ and $j$ is determined by a linear weighted ($w_{(\cdot)}$) sum of the corresponding path's properties

$$c_{i,j} = s_{i,j}(w_{\text{len}} + w_{\kappa,\text{av}}\bar{\kappa}_{i,j}^2 \\ + w_{\kappa,\text{range}}(\Delta\kappa_{i,j})^2 + w_{\text{rl}}|d_{\text{lat},j}|). \quad (4)$$

Since the curvature of a path is directly coupled to the executable speed along that path[1], we consider the average curvature $\bar{\kappa}_{i,j}$ as well as the range of curvature $\Delta\kappa_{i,j}$ (subtraction of maximum and minimum value) for the race scenario. The average measure of the curvature is built upon the absolute value of the curvature. Furthermore, the lateral offset $d_{\text{lat},j}$ of the spline segment's end point $j$ to the race line is penalized by a dedicated cost term. All cost terms are scaled with the length $s_{i,j}$ of the path segment in order to allow direct relation among the candidates. For example, the summed curvature cost of two consecutive short path segments should be same as one of a single long element.

After applying all the steps mentioned above for an entire lap on a race track, one obtains a track discretized in layers along the station $s$ and several nodes per layer in lateral direction (Figure 4).

### B. Online Part

During online execution, the offline-generated graph is further progressed in order to generate an action set. First, a local node template is extracted from the full graph (Section IV-B.1). Second, a subset of the local node template is used for each action primitive (Section IV-B.2). Third, a shortest path search is triggered in each of the action templates (Section IV-B.3). Finally, each path is post-processed by calculating a $C2$ continuous spline (Section IV-B.4) and a velocity profile (Section IV-B.5).

*1) Local node template:* The local node template specifies the nodes and corresponding edges of the full graph to be considered for the graph search in the current time step. Therefore, only the layers starting at a station $s$ in front of the ego vehicle and ending at a parameterizable planning horizon are examined further.

Since the offline part only models the static part of the race track, any kind of obstacle needs to be addressed in the online part. Therefore, the offline generated graph is manipulated by removal of occupied edges or nodes. The removal or filtering of edges instead of an infinity weight brings the advantage of fast blocking detection, faster search and avoidance of duplicate collision checks, when processing multiple objects in the same region.

Removing a node automatically removes all connected edges. Collision checking with a single node can be achieved with a fraction of the computational expenses compared to collision checking with an edge (sequence of coordinates). In light of this fact, we use node-based removal when dealing with path-like structures (e.g. vehicle predictions or blocked areas/zones) and edge-based removal when dealing with spot-like structures (e.g. a static obstacle or vehicle). All objects are represented in the form of a single or a sequence of circles for fast collision checking based on squared euclidean distances.

*2) Action templates:* In order to generate an action set consisting of multiple action primitives, the local node template is manipulated in different ways. We focus on three major primitives:

- "straight" - execute the cost minimal path in a static environment or stay behind a lead vehicle
- "left" - overtake a lead vehicle on the left
- "right" - overtake a lead vehicle on the right

Each action primitive may host one or multiple trajectories, varying in the path or velocity profile. The straight action primitive is generated by triggering the graph search without considering any dynamic obstacle. Therefore, the resulting cost minimal path normally coincides with the globally optimal race line. If there is an dynamic object in the planning horizon, it is taken into account in the velocity profile (Section IV-B.5). The overtake behavior is generated by removing the nodes to the respective left and right of the dynamic obstacle (in addition to removed entities by the local node template). That way, the graph search results in a path bypassing the lead vehicle to the left and right, respectively.

*3) Graph search:* Finding the appropriate path can be interpreted as a shortest-path problem between a start $n_i \in L_{l,\text{s}}$ and goal $n_i \in L_{l,\text{e}}$ node within the lattice. The cost optimal solution is given by the sequence of edges or nodes along the graph, which satisfies:

$$\underset{n_i \in L_l, n_j \in L_{l+1}}{\operatorname{argmin}} \sum_{l=l_{\text{s}}}^{l_{\text{e}}-1} c_{i,j}. \quad (5)$$

The goal layer is selected based on a desired planning distance ahead of the current position, which is measured along the $s$-coordinate. We aim for the path to end in a node residing on the race line at the destination layer. Since this node may be occupied with obstacles, the goal is to find the next cost optimal node in the goal layer. This is achieved by introducing one virtual node per layer, which in turn is connected to every node in the corresponding layer. Virtual edges are used for the graph search only and do not belong to the resulting drivable path. The cost associated with each of the virtual edges is weighted according to their offset to the race line.

---

[1]At the execution limit (max. lateral acceleration $a_{\text{lat}}$), the velocity $v$ is directly coupled to the curvature $\kappa$: $a_{\text{lat}} = \kappa v^2$.

*4) $C_2$ continuous spline:* As stated in Section IV-A.2, the offline computed splines do not serve $C_2$ continuity. Therefore, the nodes returned by the graph search are used to sample $C_2$ continuous splines. Similar to the offline part, the $x$ and $y$ course of the $i$-th spline segment in a sequence of $N$ splines is described by a cubic spline

$$p_i(\mu) = a_{0,i}\mu^3 + a_{0,i}\mu^2 + a_{0,i}\mu + a_{0,i}. \quad (6)$$

In addition to the start and end position constraint of each spline segment, the $C_1$ and $C_2$ constraint of two adjacent splines is considered in the boundary conditions. The start and end points of the spline chain are constrained by the heading as shown below:

$$\begin{aligned}
p_{s,i} &= a_{0,i} \\
p_{e,i} &= a_{3,i} + a_{2,i} + a_{1,i} + a_{0,i} \\
p'_{e,i} &= p'_{s,i+1} \Longleftrightarrow 3a_{3,i} + 2a_{2,i} + a_{1,i} = a_{1,i+1} \\
p''_{e,i} &= p''_{s,i+1} \Longleftrightarrow 6a_{3,i} + 2a_{2,i} = a_{2,i+1} \\
p'_{s,1} &= s_{\text{len},1} f_{\text{cs}}(\theta_{s,1}) = a_{1,1} \\
p'_{e,N} &= s_{\text{len},N} f_{\text{cs}}(\theta_{e,N}) = 3a_{3,N} + 2a_{2,N} + a_{1,N}.
\end{aligned} \quad (7)$$

*5) Velocity planner:* The velocity for every path candidate is determined based on the curvature and the corresponding local friction coefficients. The maximum executable velocity is calculated based on a forward-backwards solver, maxing out the combined acceleration potential, as described by Heilmeier et al. [15]. The start velocity is given by the planned ego velocity at the predicted (calculation time) ego position, and the goal velocity is derived from the race line's velocity profile.

When following a lead vehicle, the target velocity is altered based on a PD control law acting on the object distance, while still considering the friction limits.

## V. Experimental Results

The described method was evaluated on an electrically powered, autonomous race vehicle, called *DevBot 2.0* (Figure 1). It is equipped with various sensors and two control units. Thereby, the trajectory planner runs on an Nvidia Drive PX2 control unit. A detailed elaboration on the hardware and overall software structure can be found in Betz et al. [2], [16].

The planner runs with an average update rate of 16.8 Hz on an ARM Cortex-A57 (29.9 Hz on an Intel i7 2.9 GHz), providing one action set to the behavior planner. Each action primitive holds coordinate, heading, curvature, velocity, and acceleration commands. The planner was evaluated on the DevBot up to a speed of 212 km/h. The key parameters controlling the size of the search lattice and specifying the weights are given in Table I.

We evaluated various scenarios demonstrating the range of abilities of the planner. Figure 5a shows the avoidance of virtual static obstacles placed on the race track. The vehicle approaches the first corner with a top speed of 150 km/h. Just before entering the corner it decelerates with maximum possible acceleration and allowed lateral forces in order to achieve a fast race behavior. On the back straight, it accelerates to maximum possible speed while smoothly avoiding the obstacles placed on the track.

TABLE I: Parameters of the Offline Graph

| Parameter | Value |
|---|---|
| Lateral node separation | 0.5 m |
| Longitudinal layer separation (straights) | 30 m |
| Longitudinal layer separation (curves) | 6 m |
| Minimum planning horizon | 200 m |
| $w_{\text{len}}$ | 0 |
| $w_{\kappa,\text{av}}$ | 7500 |
| $w_{\kappa,\text{range}}$ | 15000 |
| $w_{\text{rl}}$ | 5 |

Figure 5b demonstrates an overtaking scenario executed in simulation. In this scenario, any "left" or "right" primitive is chosen whenever available. The vehicle follows the lead vehicle in a predefined safe distance around the corner, as no overtaking maneuver is possible. Once the straight is reached, the vehicle overtakes the lead vehicle and merges back onto the race line. Within this scenario a simple constant velocity (CV) prediction of 200 ms is used to anticipate the motion of the lead vehicle.

Figure 5c demonstrates a safe real-world overtaking maneuver incorporating two DevBots, both equipped with the proposed planner. The Roborace event launched in this season hosts a triggered overtaking regulation. Thereby, once the chase vehicle closes up to a predefined temporal distance (similar to the DRS regulation in Formula 1), the lead vehicle has to stay on one specified side of the track after the next curve, allowing the chase vehicle to overtake. Once the lead vehicle (blue) enters the highlighted overtaking zone, the orange vehicle accelerates and passes the vehicle. Blockage of the predefined zones is achieved by a local node template.

## VI. Discussion and Conclusion

Our planner is able to generate an action set in realtime for any race scenario, including non-convex problems. The planner was evaluated on a real race vehicle and performs well on the target hardware. The vehicle is able to smoothly avoid obstacles on the track and follow a lead vehicle whenever overtaking is not feasible. Thereby, a far planning horizon of 200 m enables recursive feasibility. The underlying global race line and curvature-oriented velocity profile contribute to a fast race execution.

Collision checking is one of the most computationally demanding tasks in our framework. Therefore, insertion of further objects in the planning horizon will lead to a reduction of the update rate. Parallelization of the collision checking task might decouple the computation time from the number of objects in the scene. Further research should also investigate the incorporation of a sophisticated dynamic prediction of other traffic participants. In order to plan proper overtake maneuvers, the consideration of the future behavior of other dynamic objects is essential.

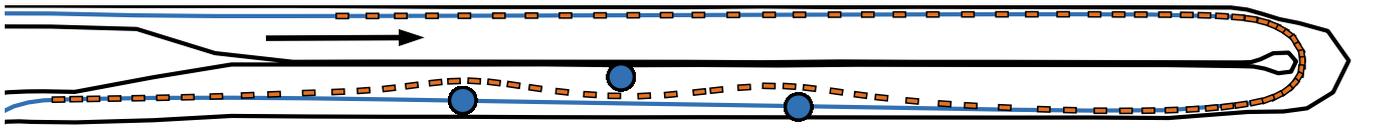

(a) *Static obstacle avoidance in simulation.* Demonstration of smooth static obstacle avoidance of a vehicle limited to 150 km/h. The vehicle (orange) slows down slightly in order to manage the evasive maneuvers and merges back into the race line.

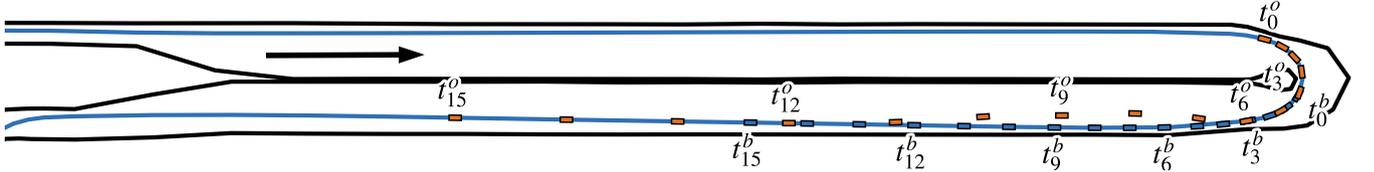

(b) *Overtaking maneuver in simulation.* The blue vehicle ($t_i^b$) is limited to 75 km/h, while the orange vehicle ($t_i^o$) is allowed to drive 150 km/h. Since there is no solution to overtake in the curve, the vehicle stays behind the lead vehicle and overtakes on entering the straight passage. Once it passes the vehicle, it merges back into the race line.

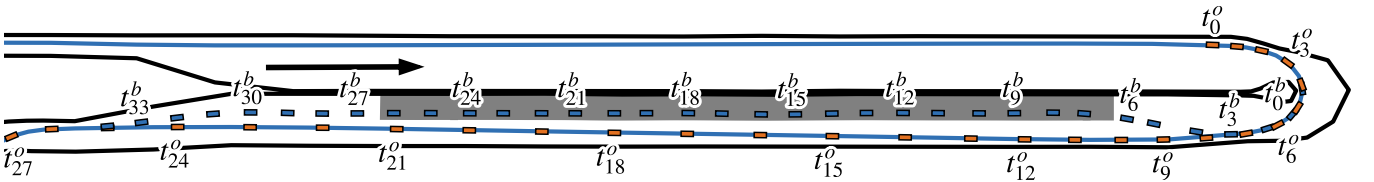

(c) *Real-world-triggered overtaking maneuver executed with two DevBots.* The blue DevBot ($t_i^b$) was limited to a top speed of 50 km/h, while the orange DevBot ($t_i^o$) is allowed to drive 100 km/h. The faster DevBot keeps a safe distance to the lead vehicle until overtaking is allowed (other vehicle is in the dedicated zone, marked in gray).

Fig. 5: Planner experiments on a straight segment of a race track laid out on an abandoned airfield. The straight segment holds a length of 500 m one way and a track width of 20 m. The globally optimal race line is highlighted in blue. Samples of the vehicle's positions are spaced at 0.3 s in (a) and 1.0 s in (b) and (c).

### Acknowledgment and contributions

Research was supported by TÜV Süd. Tim Stahl as the first author initiated the idea of this paper and is responsible for the presented concept and implementation. Alexander Wischnewski and Johannes Betz contributed to the overall concept and system design. Markus Lienkamp contributed to the conception of the research project and revised the paper critically for important intellectual content. He gave final approval of the version to be published and agrees to all aspects of the work. As guarantor, he accepts responsibility for the overall integrity of the paper.